\documentclass[10pt,letterpaper]{article}

\usepackage{cogsci}

\cogscifinalcopy % Uncomment this line for the final submission 

\usepackage{pslatex}
\usepackage{apacite}
\usepackage{graphicx} % Figures & subfigures
\usepackage{subcaption} % Figures & subfigures
\usepackage{amsmath} % Equations
\usepackage{amssymb} % Equations
\usepackage{natbib}
\usepackage{bm}
\usepackage[font=small]{caption} % Smaller caption text
\renewcommand\bibliographytypesize{\small} % Smaller bibliography text
\DeclareMathOperator*{\argmax}{arg\,max}

\usepackage{soul}
\usepackage{color}
\usepackage{xcolor,colortbl}
\definecolor{citrus}{HTML}{FAEE96}
\definecolor{orange}{HTML}{F5D29D}

\title{Learning a smooth kernel regularizer for convolutional neural networks}
 
\author{{\large \bf Reuben Feinman (reuben.feinman@nyu.edu)} \\
  Center for Neural Science \\
  New York University
  \And {\large \bf Brenden M. Lake (brenden@nyu.edu)} \\
  Department of Psychology and Center for Data Science \\
  New York University}

\begin{document}
\maketitle

\begin{abstract}
    Modern deep neural networks require a tremendous amount of data to train, often needing hundreds or thousands of labeled examples to learn an effective representation.
For these networks to work with less data, more structure must be built into their architectures or learned from previous experience. 
The learned weights of convolutional neural networks (CNNs) trained on large datasets for object recognition contain a substantial amount of structure.
These representations have parallels to simple cells in the primary visual cortex, where receptive fields are smooth and contain many regularities. 
Incorporating smoothness constraints over the kernel weights of modern CNN architectures is a promising way to improve their sample complexity. 
%We propose a smooth kernel regularizer, denoted ``SK-reg," that enforces smooth, structured convolution kernels. 
We propose a smooth kernel regularizer that encourages spatial correlations in convolution kernel weights. 
The correlation parameters of this regularizer are learned from previous experience, yielding a method with a hierarchical Bayesian interpretation.
%We show that a kernel regularizer that enforces smoothness can help constrain models for visual recognition, improving over an L2 regularization baseline.
We show that our correlated regularizer can help constrain models for visual recognition, improving over an L2 regularization baseline.

\end{abstract}

\section{Introduction}
Convolutional neural networks (CNNs) are powerful feed-forward architectures inspired by mammalian visual processing capable of learning complex visual representations from raw image data \citep{LeCun2015}.
 %\brenden{(this topic sentence is a little too generic)}
These networks achieve human-level performance in some visual recognition tasks; however, their performance often comes at the cost of hundreds or thousands of labelled examples. 
In contrast, children can learn to recognize new concepts from just one or a few examples \citep{Bloom2000, Xu2007}, evidencing the use of rich structural constraints \citep{Lake2017}.
By enforcing structure on neural networks to account for the regularities of visual data, it may be possible to substantially reduce the number of training examples they need to generalize.
In this paper, we introduce a soft architectural constraint for CNNs that enforces smooth, correlated structure on their convolution kernels through transfer learning.\footnote{Experiments from this paper can be reproduced with the code found at \url{https://github.com/rfeinman/SK-regularization}.}
We see this as an important step towards a general, off-the-shelf CNN regularizer that operates independently of previous experience.

The basis for our constraint is the idea that the weights of a convolutional kernel should in general be well-structured and smooth. 
The weight kernels of CNNs that have been trained on the large-scale ImageNet object recognition task contain a substantial amount of structure. 
These kernels have parallels to simple cells in primary visual cortex, where smooth receptive fields implement bandpass oriented filters of various scale \citep{Jones1987}.

The consistencies of visual receptive fields are explained by the regularities of image data. 
Locations within the kernel window have parallels to locations in image space, and images are generally smooth \citep{Li2009}. 
Consequently, smooth, structured receptive fields are necessary to capture important visual features like edges.
In landmark work, \cite{HubelWiesel1962} discovered edge-detecting features in the primary visual cortex of cat. 
Since then, the community has successfully modeled receptive fields in early areas of mammalian visual cortex using Gabor kernels \citep{Jones1987}.
These kernels are smooth and contain many spatial correlations.
In later stages of visual processing, locations of kernel space continue to parallel image space; however, inputs to these kernels are visual features, such as edges. 
Like earlier layers, these layers also benefit from smooth, structured kernels that capture correlations across the input space. 
\cite{Geisler2001} showed that human contour perception--an important component of object recognition--is well-explained by a model of edge co-occurrences, suggesting that correlated receptive fields are useful in higher layers of processing as well.

\begin{figure}[t]
		% top (1 subfig)
    	\begin{subfigure}[b]{\columnwidth}
            \includegraphics[width=\columnwidth]{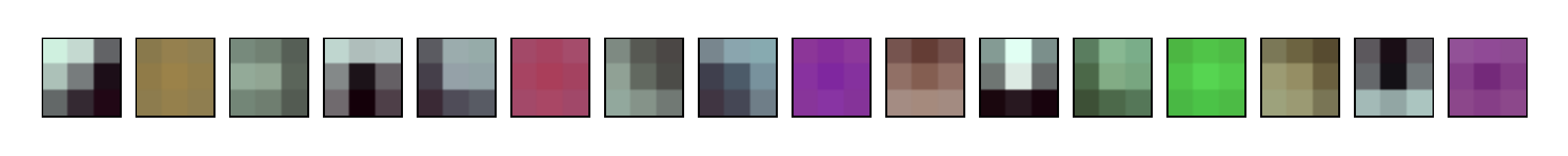}
        	\caption{VGG16 layer-1 kernels}
            \label{fig:vgg_kernels}
        \end{subfigure}
        % subfigure (b)
        \begin{subfigure}[b]{\columnwidth}
            \centering
            \begin{subfigure}[b]{0.49\hsize}
                \centering
                \includegraphics[width=0.9\hsize]{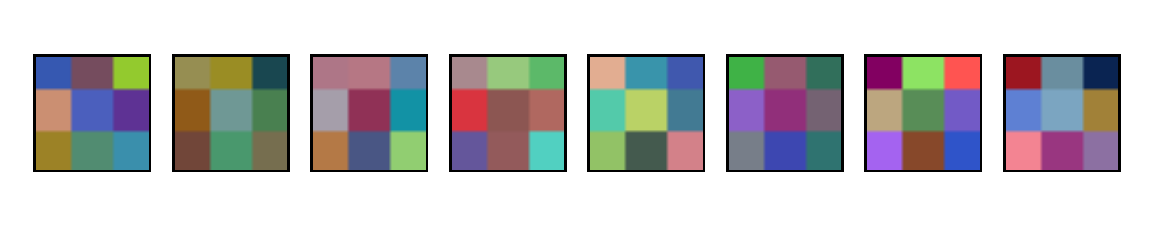}
                \caption{i.i.d. Gaussian (L2-reg)}
                \label{fig:ig_samples}
            \end{subfigure}
            \centering
            \begin{subfigure}[b]{0.49\hsize}
                \centering
                \includegraphics[width=0.9\hsize]{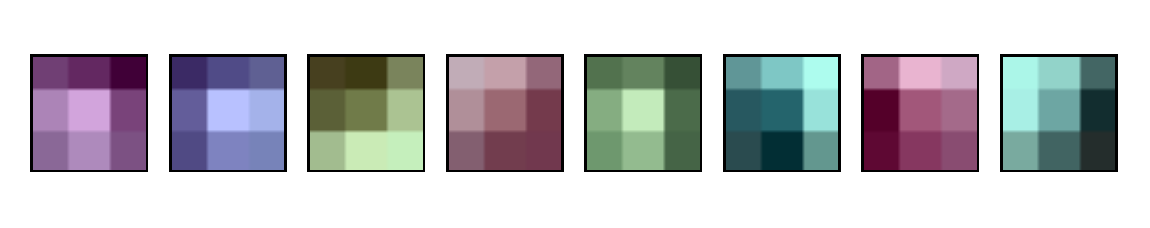}
                \caption{correlated Gaussian (SK-reg)}
                \label{fig:cg_samples}
            \end{subfigure}
        \end{subfigure}
    \caption{Kernel priors for VGG16. The layer-1 convolution kernels of VGG16, shown in (a), possess considerable correlation structure. An i.i.d. Gaussian prior that has been fit to the VGG layer-1 kernels, samples from which are shown in (b), captures little of the structure in these kernels. A correlated multivariate Gaussian prior, samples from which are shown in (c), captures the correlation structure of these kernels well.}
    \label{fig:kernel_viz}
\end{figure}

Despite the clear advantages of structured receptive fields, constraints placed on the convolution kernels of CNNs are typically chosen to be as general as possible, with neglect of this structure. 
L2 regularization--the standard soft constraint applied to kernel weights, which is interpreted as a zero-mean, independent identically distributed (i.i.d.) Gaussian prior--treats each weight as an independent random variable, with no correlations between weights expected a priori. 
Fig. \ref{fig:kernel_viz} shows the layer-1 convolutional kernels of VGG16, a ConvNet trained on the large-scaled ImageNet object recognition task \citep{vgg16}. 
Fig. \ref{fig:ig_samples} shows some samples from an i.i.d. Gaussian prior, the equivalent of L2 regularization. 
Clearly, this prior captures little of the correlation structure possessed by the kernels. 

A simple and logical extension of the i.i.d. Gaussian prior is a correlated multivariate Gaussian prior, which is capable of capturing some of the covariance structure in the convolution kernels. 
Fig. \ref{fig:cg_samples} shows some samples from a correlated Gaussian prior that has been fit to the VGG16 kernels. 
This prior provides a much better model of the kernel distribution. 
In this paper, we perform a series of controlled CNN learning experiments using a smooth kernel regularizer--which we denote ``SK-reg"--based on a correlated Gaussian prior. 
The correlation parameters of this prior are obtained by fitting a Gaussian to the learned kernels from previous experience. 
%We formulate this algorithm as a method for probabilistic inference in a hierarchical Bayesian model.
We compare SK-reg to standard L2 regularization in two object recognition use cases: one with simple silhouette images, and another with Tiny ImageNet natural images. 
In the condition of limited training data, SK-reg yields improved generalization performance.

\section{Background}
Our goal in this paper is to introduce new a priori structure into CNN receptive fields to account for the regularities of image data and help reduce the sample complexity of these models. Previous methods from this literature often require a fixed model architecture that cannot be adjusted from task to task. 
In contrast, our method enforces structure via a statistical prior over receptive field weights, allowing for flexible architecture adaption to the task at hand.
Nevertheless, in this section we review the most common approaches to structured vision models.

A popular method to enforce structure on visual recognition models is to apply a fixed, pre-specified representation. 
In computational vision, models of image recognition consist of a hierarchy of transformations motivated by principles from neuroscience and signal processing  \citep[e.g.,][]{Serre2007, Bruna2013}. 
These models are effective at extracting important statistical features from natural images, and they have been shown to provide a useful image representation for SVMs, logistic regression and other ``shallow" classifiers when applied to recognition tasks with limited training data. 
Unlike CNNs, the kernel parameters of these models are not learned by gradient descent. 
As result, these features may not be well-adapted to the specific task at hand.

In machine learning, it is commonplace to use the features from CNNs trained on large object recognition datasets as a generic image representation for novel vision tasks \citep{Donahue2014, Razavian2014}.
Due to the large variety of training examples that these CNNs receive, the learned features of these networks provide an effective representation for a range of new recognition tasks. Some \textit{meta-learning} algorithms use a similar form of feature transfer, where a feature representation is first learned via a series of classification episodes, each with a different support set of classes \citep[e.g.,][]{Vinyals2016}. As with pre-specified feature models, the representations of these feature transfer models are fixed for the new task; thus, performance on the new task may be sub-optimal.

Beyond fixed feature representations, other approaches use a pre-trained CNN as an initialization point for a new network, following with a fine-tuning phase where network weights are further optimized for a new task via gradient descent \citep[e.g.,][]{Girshick2014, Girshick2015}.
By adapting the CNN representation to the new task, this approach often enables better performance than fixed feature methods; however, when the scale of the required adaptation is large and the training data is limited, fine-tuning can be difficult.
\cite{Finn2017} proposed a modification of the pre-train/fine-tune paradigm called model-agnostic meta-learning (MAML) that enables flexible adaptation in the fine-tuning phase when the training data is limited.
During pre-training (or \textit{meta-learning}), MAML optimizes for a representation that can be easily adapted to a new learning task in a later phase.
Although effective for many use cases, this approach is unlikely to generalize well when the type of adaptation required differs significantly from the adaptations seen in the meta-learning episodes.
A shared concern for all pre-train/fine-tune methods is that they require a fixed model architecture between the pre-train and fine-tune phases.

The objective of our method is distinct from those of fixed feature representations and pre-train/fine-tune algorithms. 
In this paper, we study the structure in the learned parameters of vision models, with the aim of extracting general structural principles that can be incorporated into new models across a broad range of learning tasks.
SK-reg serves as a parameter prior over the convolution kernels of CNNs and has a theoretical foundation in Bayesian parameter estimation. 
This approach facilitates a CNN architecture and representation that is adapted to the specific task at hand, yet that possesses adequate structure to account for the regularities of image data.
The SK-reg prior is learned from previous experience, yielding an interpretation of our algorithm as a method for hierarchical Bayesian inference. 

Independently of our work, \cite{Atanov2019} developed the \textit{deep weight prior}, an algorithm to learn and apply a CNN kernel prior in a Bayesian framework. 
Unlike our prior, which is parameterized by a simple multivariate Gaussian, the deep weight prior uses a sophisticated density estimator parameterized by a neural network to model the learned kernels of previously-trained CNNs.
%The complexity of this estimator raises concerns about its ability to generalize.
The application of this prior to new learning tasks requires variational inference with a well-calibrated variational distribution.
Our goal with SK-reg differs in that we aim to provide an interpretable, generalizable prior for CNN weight kernels that can be applied to existing CNN training algorithms with little modification. 

\section{Bayesian interpretation of regularization}
%\brenden{(perhaps rename section header as "Empirical Bayesian regularization" or hierarchical Bayesian interpretation...)} 
%\reuben{Don't think it makes sense to change this section's name. Ok if I simply add a new sub-section about hierarchical Bayes?}
\begin{figure*}[t]
    \centering
    \includegraphics[width=\hsize]{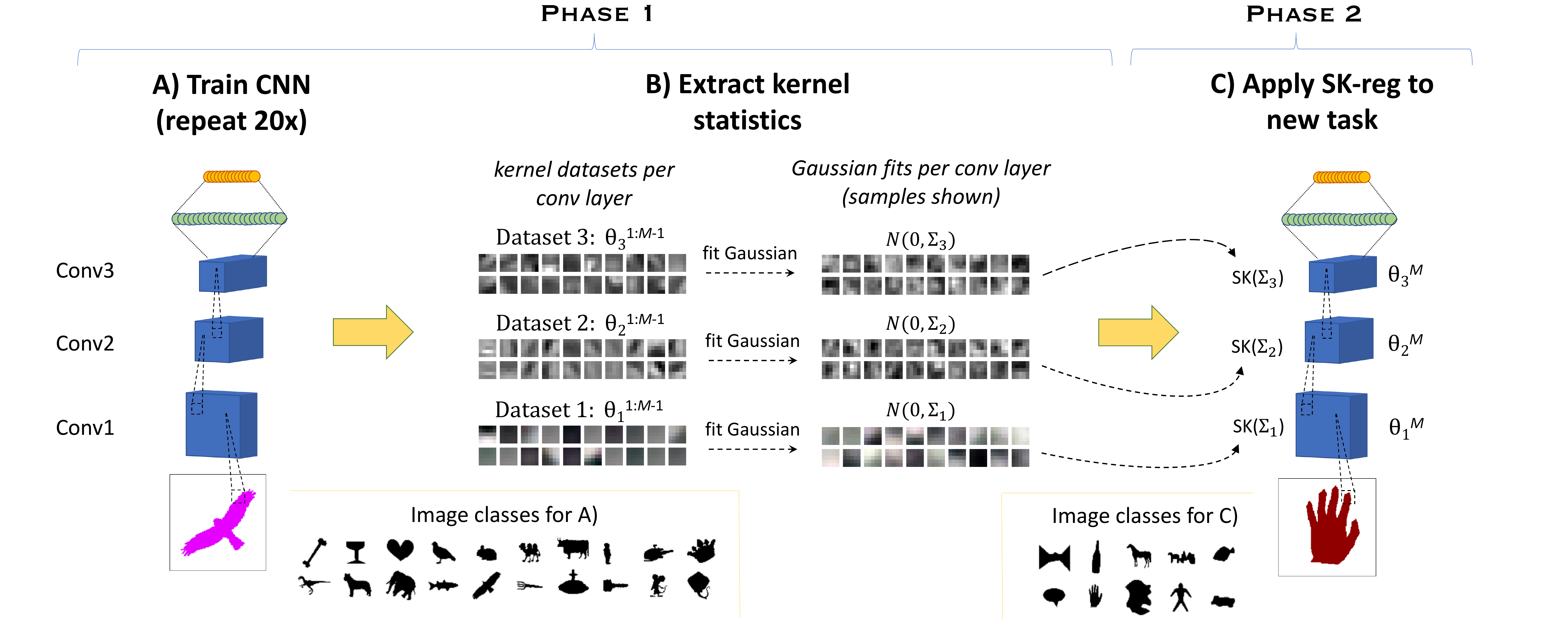}
	\caption{SK-reg workflow. A) First, a CNN is trained repeatedly (20x) on an object recognition task. B) Next, the learned parameters of each CNN are studied and statistics are extracted. For each convolution layer, kernels from the multiple CNNs are consolidated, yielding a kernel dataset for the layer. A multivariate Gaussian is fit to each kernel dataset. C) SK-reg is applied to a fresh CNN trained on a new learning task with limited training data (possibly with a different architecture or numbers of kernels), using the resulting Gaussians from each layer.}
    \label{fig:phases_explanation}
\end{figure*}

From the perspective of Bayesian parameter estimation, the L2 regularization objective can be interpreted as performing \textit{maximum a-posteriori} inference over CNN parameters with a zero-mean, i.i.d. Gaussian prior. Here, we review this connection, and we discuss the extension to SK-reg.

\subsubsection{L2 regularization.}
Assume we have a dataset \(X = \{x_1,...,x_N\}\) and \(Y = \{y_1,...,y_N\}\) consisting of \(N\) images \(x_i\) and \(N\) class labels \(y_i\). 
Let \(\theta\) define the parameters of the CNN that we wish to estimate. 
The L2 regularization objective is stated as follows:
\begin{align}
    \tilde{\theta} = \argmax_{\theta} 
    \text{ log } p(Y \mid \theta; X) - \lambda * \theta^T \theta.
    \label{eq:l2_obj}
\end{align}
Here, the first term of our objective is our prediction accuracy (classification log-likelihood), and the second term is our L2 regularization penalty. 

From a Bayesian perspective, this objective can be thought of as finding the \textit{maximum a-posteriori} (MAP) estimate of the network parameter posterior $p(\theta \mid Y; X) \propto p(Y \mid \theta; X)*p(\theta)$, leading to the optimization problem
\begin{align}
    \tilde{\theta} = \argmax_{\theta} \text{ log } p(Y \mid \theta; X) + \text{log } p(\theta).
\label{eq:map_obj}
\end{align}
To make the connection with  L2 regularization, we assume a zero-mean, i.i.d Gaussian prior over the parameters $\theta$ of a weight kernel, written as
\begin{align}
    p(\theta) = \frac{1}{Z} \text{exp} 
    \big( -\frac{1}{2\sigma^2}\theta^T \theta \big).
    \label{eq:iid_gaussian}
\end{align}
With this prior, Eq. \ref{eq:map_obj} becomes 
\begin{align*}
    \tilde{\theta} = \argmax_{\theta} 
    \text{ log } p(Y \mid \theta; X) - \frac{1}{2\sigma^2}\theta^T \theta,
\end{align*}
which is the L2 objective of Eq. \ref{eq:l2_obj}, with $\lambda = \frac{1}{2\sigma^2}$.

\subsubsection{SK regularization.}
The key idea behind SK-reg is to extend the L2 Gaussian prior to include a non-diagonal covariance matrix; i.e., to add correlation. In the case of SK-reg, the prior over kernel weights $\theta$ of Eq. \ref{eq:iid_gaussian} becomes
\begin{align*}
    p(\theta) = \frac{1}{Z} \text{exp} 
    \big( -\frac{1}{2} \theta^T \Sigma^{-1} \theta \big)
\end{align*}
for some covariance matrix $\Sigma$, and the new objective is written
\begin{align}
    \tilde{\theta} = \argmax_{\theta} 
    \text{ log } p(Y \mid \theta; X) - \lambda * \theta^T \Sigma^{-1} \theta.
    \label{eq:sk_obj}
\end{align}

\subsubsection{Hierarchical Bayes.}
\begin{figure}[h!]
    \centering
    \includegraphics[width=\columnwidth]{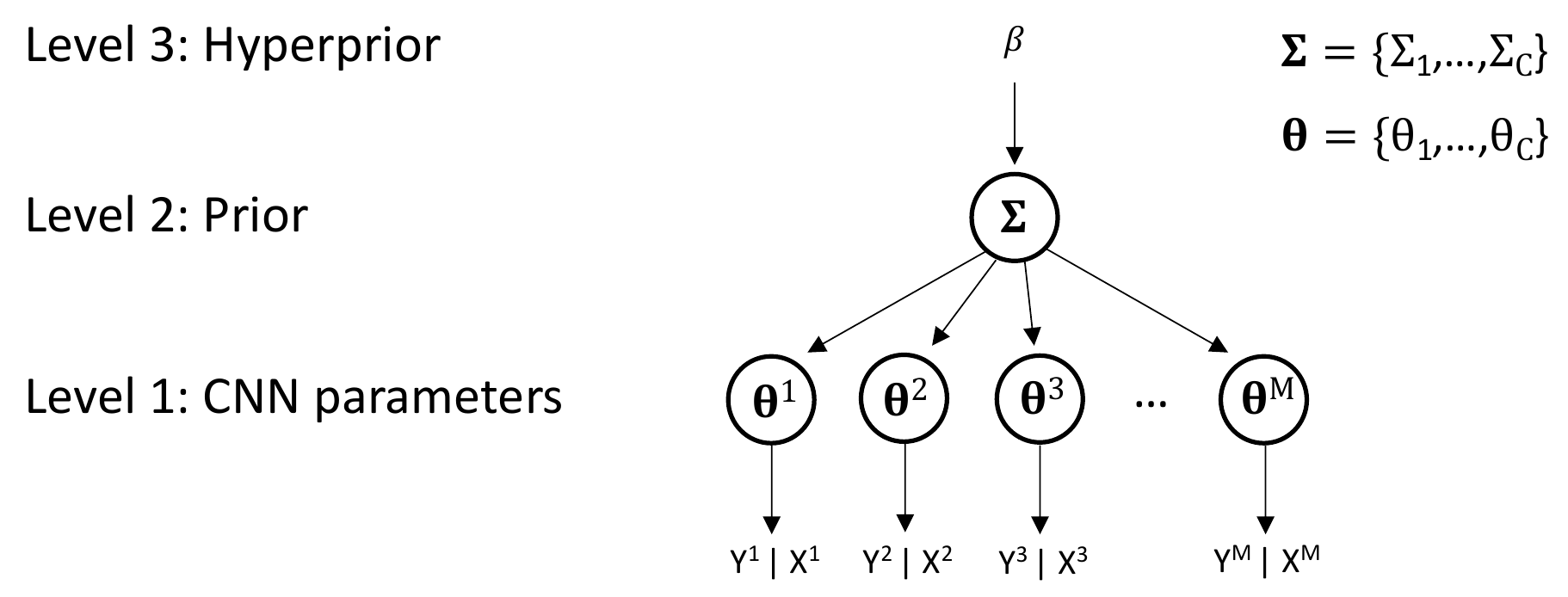}
	\caption{A hierarchical Bayesian interpretation of SK-reg. A point estimate of prior parameters $\bm{\Sigma}$ is first computed with MAP estimation. Next, this prior is applied to estimate CNN parameters $\bm{\theta}^j$ in a new task.}
    \label{fig:hbm_figure}
\end{figure}

When $\Sigma$ is learned from previous experience, SK-reg can be interpreted as approximate inference in a hierarchical Bayesian model. 
The SK regularizer for a CNN with $C$ layers, $\bm{\Sigma}=\{\Sigma_1,\dots,\Sigma_C\}$, assumes a unique zero-mean Gaussian prior $\mathcal{N}(\theta_i;0,\Sigma_i)$ over the weight kernels for each convolutional layer, $\bm{\theta}=\{\theta_1,\dots,\theta_C\}$.
Due to the regularities of the visual world, it is plausible that effective general priors exist for each layer of visual processing. 
In this paper, transfer learning is used to fit the prior covariances $\bm{\Sigma}$ from previous datasets $X^{1:M-1}$ and $Y^{1:M-1}$, which informs the solution for a new problem $X^{M}$ and $Y^{M}$, yielding the hierarchical Bayesian interpretation depicted in Fig. \ref{fig:hbm_figure}.
Task-specific CNN parameters $\bm{\theta}^{1:M}$ are drawn from a common $\bm{\Sigma}$, and  $\bm{\Sigma}$ has a hyperprior specified by $\beta$. Ideal inference would compute $p(Y^{M}|Y^{1:M-1};X^{1:M})$, marginalizing over $\bm{\theta}^{1:M}$ and $\bm{\Sigma}$.

We propose a very simple empirical Bayes procedure for learning the kernel regularizer in Eq. \ref{eq:sk_obj} from data. First, $M-1$ CNNs are fit independently to the datasets $X^{1:M-1}$ and $Y^{1:M-1}$ using standard methods, in this case optimizing Eq. \ref{eq:l2_obj} to get point estimates $\tilde{\bm{\theta}}^{1:M-1}$. Second, a point estimate $\tilde{\bm{\Sigma}}$ is computed by maximizing $p(\bm{\Sigma}|\tilde{\bm{\theta}}^{1:M-1}; \beta)$, which is a simple regularized covariance estimator. Last, for a new task $M$ with training data $X^M$ and $Y^M$, a CNN with parameters $\bm{\theta}^M$ is trained with the SK-reg objective (Eq. \ref{eq:sk_obj}), with $\bm{\Sigma} = \tilde{\bm{\Sigma}}$.

This procedure can be compared with the hierarchical Bayesian interpretation of MAML \citep{Grant2018}. Unlike MAML, our method allows flexibility to use different architectures for different datasets/episodes, and the optimizer for $\bm{\theta}^M$ is run to convergence rather than just a few steps.

\section{Experiments}
We evaluate our approach within a set of controlled visual learning environments. 
SK-reg parameters $\Sigma_i$ for each convolution layer $\theta_i$ are determined by fitting a Gaussian to the kernels acquired from an earlier learning phase.
We divide our learning tasks into two unique phases, applying the same CNN architecture in each case.
We note that our approach does not require a fixed CNN architecture across these two phases; the number of feature maps in each layer may be easily adjusted.
A depiction of the two learning phases is given in Fig. \ref{fig:phases_explanation}.

\subsubsection{Phase 1.} 
The goal of phase 1 is to extract general principles about the structure of learned convolution kernels by training an array of CNNs and collecting statistics about the resulting kernels. 
In this phase, we train a CNN architecture to classify objects using a sufficiently large training set with numerous examples per object class. 
Training is repeated multiple times with unique random seeds, and the learned convolution kernels are stored for each run. 
During this phase, standard L2 regularization is applied to enforce a minimal constraint on each layer's weights (optimization problem of Eq. \ref{eq:l2_obj}). 
After training, the convolution kernels from each run are consolidated, holding each layer separate.
A multivariate Gaussian is fit to the centered kernel dataset of each layer, yielding a distribution $N(0,\Sigma_i)$ for each convolution layer $i$.
To ensure the stability of the covariance estimators, we apply shrinkage to each covariance estimate, mixing the empirical covariance with an identity matrix of equal dimensionality. This can be interpreted as a hyperprior $p(\bm{\Sigma};\beta)$  (Fig. \ref{fig:hbm_figure}) that favors small correlations. The optimal mixing parameter is determined via cross-validation.

\subsubsection{Phase 2.} 
In phase 2, we test the aptitude of SK-reg on a new visual recognition task, applying the covariance matrices $\Sigma_i$ obtained from phase 1 to regularize each convolution layer $i$ in a freshly-trained CNN (optimization problem of Eq. \ref{eq:sk_obj}). 
In order to adequately test the generalization capability of our algorithm, we use a new set of classes that differ from the phase 1 classes in substantial ways, and we provide just a few training examples from each class. Performance of SK-reg is compared against standard L2 regularization.

\subsection{Silhouettes}
As a preliminary use case, we train our network using the binary shape image dataset developed at Brown University\footnote{The binary shape dataset is available in the ``Databases" section at \url{http://vision.lems.brown.edu}}, henceforth denoted ``Silhouettes."
Silhouette images are binary masks that depict the structural form of various object classes. Simple shape-based stimuli such as these provide a controlled learning environment for studying the inductive biases of CNNs \citep{Feinman2018}.
We select a set of 20 well-structured silhouette classes for phase 1, and a set of 10 unique, well-structured classes for phase 2 that differ from phase 1 in their consistency and form. The images are padded to a fixed size of $200 \times 200$.

\begin{figure}[t]
    \centering
    \includegraphics[width=\columnwidth]{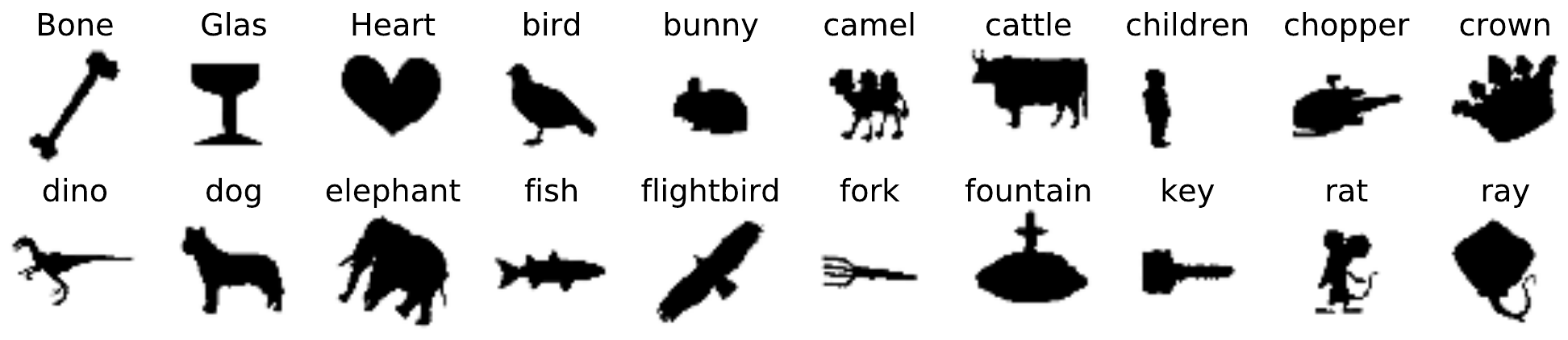}
	\caption{Exemplars of the phase 1 silhouette object classes.}
    \label{fig:sil_classes1}
\end{figure}

During phase 1, we train our network to perform 20-way object classification.
Exemplars of the phase 1 classes are shown in Fig. \ref{fig:sil_classes1}. 
The number of examples varies for each class, ranging from 12 to 49 with a mean of 24. Class weighting is used to remedy class imbalances. 
To add complexity to the silhouette images, colors are assigned randomly to each silhouette before training. 
During training, random translations, rotations and horizontal flips are applied at each training epoch to improve generalization performance.

\begin{table}[t]
    \centering
    {\small
    \begin{tabular}{lllll} 
        \hline
        Layer & Window & Stride & Features & $\lambda$\\
        \hline
        Input (200x200x3) \\
        Conv2D  & 5x5 & 2 & 5 & 0.05\\
        MaxPooling2D & 3x3 & 3 \\
        Conv2D & 5x5 & 1 & 10 & 0.05 \\
        MaxPooling2D & 3x3 & 2 \\
        Conv2D & 5x5 & 1 & 8 & 0.05\\
        MaxPooling2D & 3x3 & 1 \\
        %Dropout (0.2) \\
        FullyConnected & & & 128 & 0.01 \\
        %Dropout (0.5) \\
        Softmax \\
        \hline
    \end{tabular}
    }
    \caption{CNN architecture. Layer hyperparameters include window size, stride, feature count, and regularization weight ($\lambda$). Dropout is applied after the last pooling layer and the fully-connected layer with rates 0.2 and 0.5, respectively.}
    \label{tab:cnn_architecture} 
\end{table}

We use a CNN architecture with 3 convolution layers, each followed by a max pooling layer (see Table \ref{tab:cnn_architecture}). 
Hyperparameters including convolution window size, pool size, and filter counts were selected via randomized grid-search, using a validation set with examples from each class to score candidate values.
A rectified linear unit (ReLU) nonlinearity is applied to the output of each convolution layer, as well as to the fully-connected layer.
The network is trained 20 times using the Adam optimizer, each time with a unique random initialization. 
It achieves an average validation accuracy of 97.7\% across the 20 trials, indicating substantial generalization.

\begin{figure}[t]
    	\begin{subfigure}[b]{\columnwidth}
            \includegraphics[width=\columnwidth]{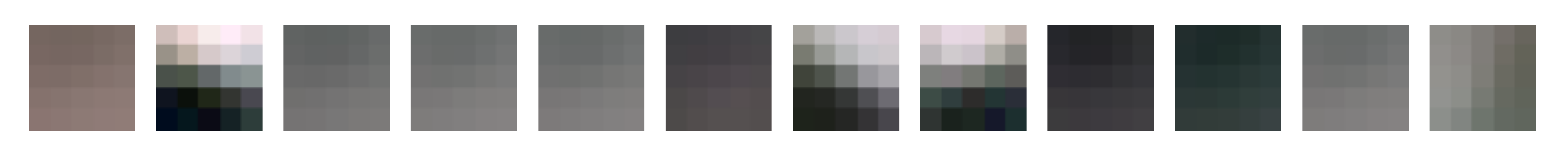}
        	\caption{First-layer kernels}
            \label{fig:sil_kernels}
        \end{subfigure}
    	\begin{subfigure}[b]{\columnwidth}
            \includegraphics[width=\columnwidth]{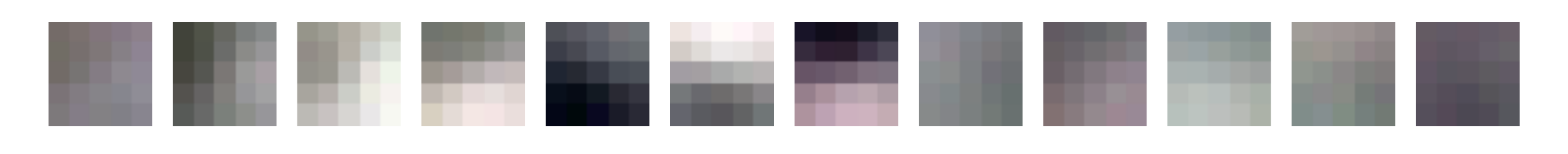}
        	\caption{Gaussian samples}
            \label{fig:sil_cg_samples}
        \end{subfigure}
    \caption{Learned first-layer kernels vs. Gaussian samples. (a) depicts some of the learned first-layer kernels acquired from phase 1 silhouette training. For comparison, (b) shows a few samples from a multivariate Gaussian that was fit to the first-layer kernel dataset.}
    \label{fig:sil_kernels_viz}
\end{figure}

Following the completion of phase 1 training, a kernel dataset is obtained for each convolution layer by consolidating the learned kernels for that layer from the 20 trials. 
Covariance matrices $\Sigma_i$ for each layer $i$ are obtained by fitting a multivariate Gaussian to the layer's kernel dataset.
For a first-layer convolution with window size $K \times K$, this Gaussian has dimensionality $3K^2$, equal to the window area times RGB depth. 
We model the input channels as separate variables in layer 1 because these channels have a consistent interpretation as the RGB color channels of the input image. 
For remaining convolution layers, where the interpretation of input channels may vary from case to case, we treat each input channel as an independent sample from a Gaussian with dimensionality $K^2$. The kernel datasets for each layer are centered to ensure zero mean, typically requiring only a small perturbation vector.

To ensure that our multivariate Gaussians model the kernel data well, we computed the cross-validated log-likelihoods of this estimator on each layer's kernel dataset and compared them to those of an i.i.d. Gaussian estimator fit to the same data. The multivariate Gaussian achieved an average score of 358.5, 413.3 and 828.1 for convolution layers 1, 2 and 3, respectively. In comparison, the i.i.d. Gaussian achieved an average score of 144.4, 289.6 and 621.9 for the same layers. These results confirm that our multivariate Gaussian provides an improved model of the kernel data. Some examples of the first-layer convolution kernels are shown in Fig. \ref{fig:sil_kernels_viz} alongside samples from our multivariate Gaussian that was fit to the first-layer kernel dataset. The samples appear structurally consistent with our phase 1 kernels.

\begin{figure}[t]
    \centering
    \includegraphics[width=\columnwidth]{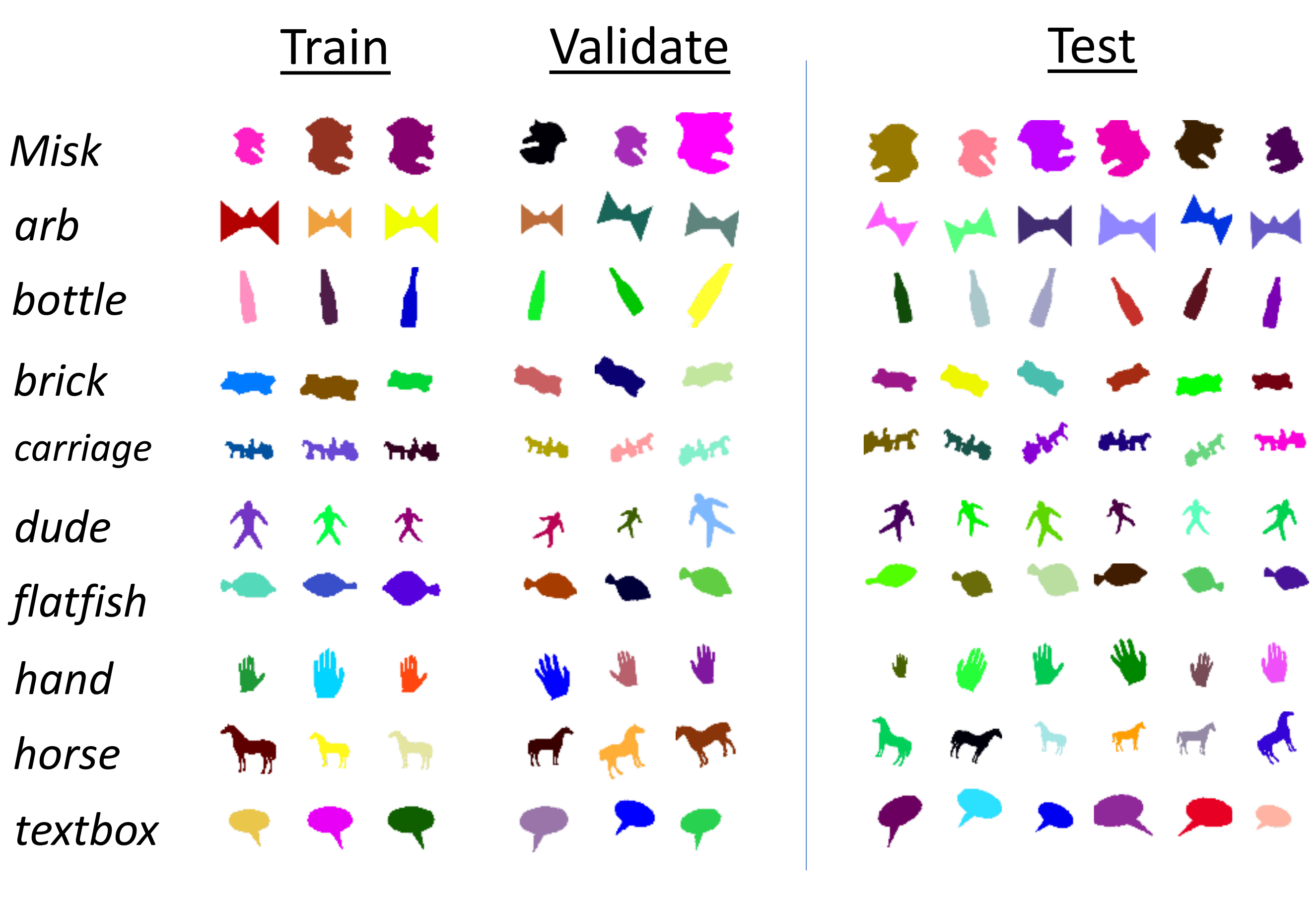}
	\caption{Silhouettes phase 2 datasets. 3 examples per class are provided in both the train and validation sets. A holdout test set with 6 examples per class is used to evaluate final model performance.}
    \label{fig:phase2_data}
\end{figure}

In phase 2, we train our CNN on a new 10-way classification task, providing the network with just 3 examples per class for gradient descent training and 3 examples per class for validation. 
Colors are again added at random to each silhouette in the dataset. 
The network is initialized randomly, and we apply SK-reg to the convolution kernels of each layer during training using the covariance matrices obtained in phase 1. 
Our validation set is used to track and save the best model over the course of the training epochs (early stopping).
A holdout set with 6 examples per class is used to assess the final performance of the model.
A depiction of the train, validation and test sets used for phase 2 is given in Fig. \ref{fig:phase2_data}.
The validation and test images have been shifted, translated and flipped to make for a more challenging generalization test.
Similar to phase 1, random shifts, rotations and horizontal flips are applied to the training images at each training epoch.  
As a baseline, we also train our CNN using standard L2 regularization.

The regularization weight $\lambda$ is an important hyperparameter of both SK and L2 regularization. 
Before performing the phase 2 training assessment, we use a validated grid search to select the optimal $\lambda$ for each regularization method, applying our train/validate sets.\footnote{To yield interpretable $\lambda$ values that can be compared between the SK and L2 cases, we normalize each covariance matrix to unit determinant by applying a scaling factor, such that det($c\Sigma$) = det($I$)}
The same weight $\lambda$ is applied to each convolution layer, as done in phase 1.

\begin{table}[t]
    \centering
    \begin{tabular}{llll} 
        \hline
        Method & $\lambda$ & Cross-entropy & Accuracy \\
        \hline
        L2 & 0.214 & 2.000 (+/- 0.033) & 0.530 (+/- 0.013) \\
        SK & 0.129 & 0.597 (+/- 0.172) & 0.821 (+/- 0.056) \\
        \hline
    \end{tabular}
    \caption{Silhouettes phase 2 results. For each regularization method, the optimal regularization weight $\lambda$ was selected via grid-search. Results show the average cross-entropy and classification accuracy achieved on the holdout test set over 10 phase 2 training runs.} 
    \label{tab:phase2_results} 
\end{table}

\subsubsection{Results.}
With our optimal $\lambda$ values selected, we trained our CNN on the 10-way phase 2 classification task of Fig. \ref{fig:phase2_data}, comparing SK regularization to a baseline L2 regularization model.
Average results for the two models collected over 10 training runs are presented in Table \ref{tab:phase2_results}.
Average test accuracy is improved by roughly ~55\% with the addition of SK reg, a substantial performance boost from 53.0\% correct to 82.1\% correct. Clearly, a priori structure is beneficial to generalization in this use case. An inspection of the learned kernels confirms that SK-reg encourages the structure we expect; these kernels look visually similar to samples from the Gaussian (e.g. Fig. \ref{fig:sil_kernels_viz}).

\subsection{Tiny ImageNet}
Our silhouette experiment demonstrates the effectiveness of SK-reg when the parameters of the regularizer are determined from the structure of CNNs trained on a similar image domain.
However, it remains unclear whether these regularization parameters can generalize to novel image domains.
Due to the nature of the silhouette images, the silhouette recognition task encourages representations with properties that are desirable for object recognition tasks in general. 
Categorizing silhouettes requires forming a rich representation of shape, and shape perception is critical to object recognition. 
Therefore, this family of representation may be useful in a variety of object recognition tasks.

\begin{figure}[t]
    \centering
    \includegraphics[width=\columnwidth]{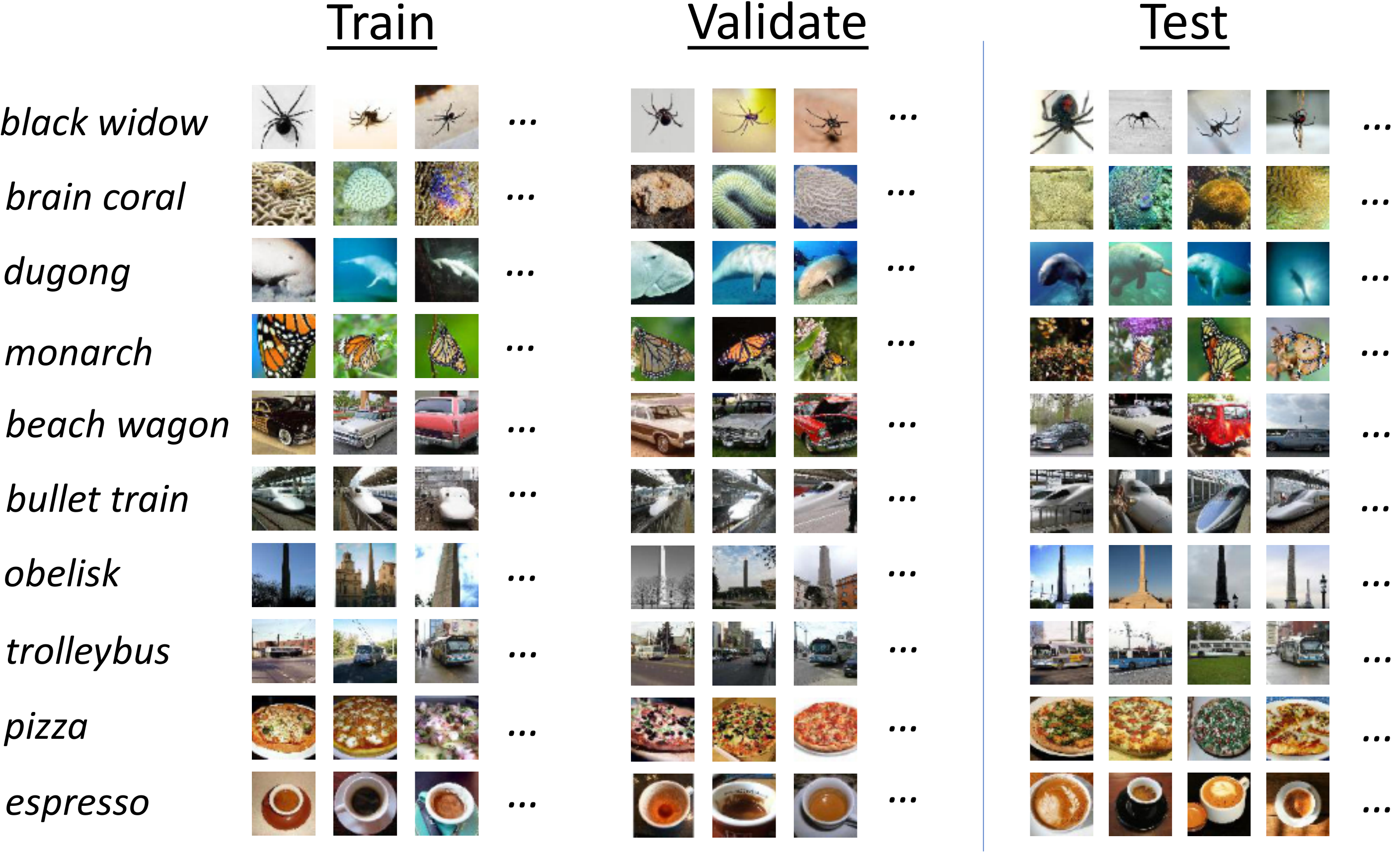}
	\caption{Tiny ImageNet datasets. 10 classes were selected to form a 10-way classification task. The train and validate sets each contain 10 examples per class. The holdout test set contains 20 examples per class.}
    \label{fig:imagenet_data}
\end{figure}

To test whether our kernel priors obtained from silhouette training generalize to a novel domain, we applied SK-reg to a simplified version of the Tiny ImageNet visual recognition challenge, using covariance parameters fitted to silhouette-trained CNNs. 
Tiny ImageNet images were up-sampled with bilinear interpolation from their original size of $64 \times 64$ to mirror the Silhouette size $200 \times 200$. 
We selected 10 well-structured ImageNet classes that contain properties consistent with the silhouette images.\footnote{Desirable classes have a uniform, centralized object with consistent shape properties and a distinct background.}
%Some examples of the 10 classes are shown in figure \ref{fig:imagenet_data}. 
We performed 10-way image classification with these classes, using the same CNN architecture from Table \ref{tab:cnn_architecture} and applying the SK-reg soft constraint.
The network is provided 10 images per class for training and 10 per class for validation.
Because of the increased complexity of the Tiny ImageNet data, a larger number of examples per class is merited to achieve good generalization performance.
%Next to the standards of the image feature learning literature, this training set size is comparatively small.
A holdout test set with 20 images per class is used to evaluate performance. 
Fig. \ref{fig:imagenet_data} shows a breakdown of the train, validate and test sets.

A few modifications were made to account for the new image data.
First, we modified the phase 1 silhouette training used to acquire our covariance parameters, this time applying random colors to both the foreground and background of each silhouette. 
Previously, each silhouette overlaid a strictly white background. Consequently, the edge detectors of the learned CNNs would be unlikely to generalize to novel color gradients. 
Second, we added additional regularization to our covariance estimators to avoid over-fitting and help improve the generalization capability of the resulting kernel priors. 
Due to the nature of the phase 2 task in this experiment, and the extent to which the images differ from phase 1,  additional regularization was necessary to ensure that our kernel priors could generalize.
Specifically, we applied L1-regularized inverse covariance estimation \citep{Friedman2008} to estimate each $\Sigma_i$, which can be interpreted as a hyperprior $p(\bm{\Sigma};\beta)$ (Fig. \ref{fig:hbm_figure}) that favors a sparse inverse covariance \citep{Lake2010}.
%\brenden{(You need to say more about why this was necessary, or why this changed was needed between the two experiments.)}

Similar to the silhouettes experiment, the validation set is used to select weighting hyperparameter $\lambda$ and to track the best model over the course of learning.
As a baseline, we again compared SK-reg to a $\lambda$-optimized L2 regularizer.

\subsubsection{Results.}
SK-reg improved the average holdout performance received from 10 training runs as compared to an L2 baseline, both in accuracy and cross-entropy. Results for each regularization method, as well as their optimal $\lambda$ values, are reported in Table \ref{tab:imagenet_results}. An improvement of 8\% in test accuracy suggests that some of the structure captured by our kernel prior is useful even in a very distinct image domain. The complexity of natural images like ImageNet is vast in comparison to simple binary shape masks; nonetheless, our prior from phase 1 silhouette training is able to influence ImageNet learning in a manner that is beneficial to generalization.

\begin{table}[t]
    \centering
    \begin{tabular}{llll} 
        \hline
        Method & $\lambda$ & Cross-entropy & Accuracy \\
        \hline
        L2 & 0.450 & 1.073 (+/- 0.102) & 0.700 (+/- 0.030) \\
        SK & 0.450 & 0.956 (+/- 0.180) & 0.776 (+/- 0.035) \\
        \hline
    \end{tabular}
    \caption{Tiny ImageNet SK-reg and L2 results. Table shows the average cross-entropy and classification accuracy achieved on the holdout test set over 10 training runs.} 
    \label{tab:imagenet_results} 
\end{table}

\section{Discussion}
Using a set of controlled visual learning experiments, our work in this paper demonstrates the potential of structured receptive field priors in CNN learning tasks.
Due to the properties of image data, smooth, structured receptive fields have many desirable properties for visual recognition models. 
In our experiments, we have shown that a simple multivariate Gaussian model can effectively capture some of the structure in the learned receptive fields of CNNs trained on simple object recognition tasks. 
Samples from the fitted Gaussians are visually consistent with learned receptive fields, and when applied as a model prior for new learning tasks, these Gaussians can help a CNN generalize in conditions of limited training data.
We demonstrated our new regularization method in two simple use cases.
Our silhouettes experiment shows that, when the parameters of SK-reg are determined from CNNs trained on a similar image domain to that of the new task, the performance increase that results in the new task can be quite substantial--as large as 55\% over an L2 baseline. 
Our Tiny ImageNet experiment demonstrates that SK-reg is capable of encoding generalizable structural principles about the correlations in receptive fields; the statistics of learned parameters in one domain can be useful in a completely new domain with substantial differences.

The Gaussians that we fit to kernel data in phase 1 of our experiments could be overfit to the CNN training runs. 
We have discussed the application of sparse inverse covariance (precision) estimation as one approach to reduce over-fitting. 
In future work, we would like to explore a Gaussian model with graphical connectivity that is specified by a 2D grid MRF. 
Model fitting would consist of optimizing the non-zero precision matrix values subject to this pre-specified sparsity. 
The grid MRF model is enticing for its potential to serve as a general ``smoothness" prior for CNN receptive fields. 
Ultimately, we hope to develop a general-purpose kernel regularizer that does not depend on transfer learning.
%\brenden{(Good place to point out future directions of finding other ways to enforce smoothness that do not depend on transfer learning.)}

Although a Gaussian can model some kernel families sufficiently, other families would give it a difficult time. The first-layer kernels of AlexNet--which are $11 \times 11$ and are visually similar to Gabor wavelets and derivative kernels--are not well-modeled by a multivariate Gaussian. A more sophisticated prior is needed to model kernels of this size effectively. In future work, we hope to investigate more complex families of priors that can capture the regularities of filters such as Gabors and derivatives. Nevertheless, a simple Gaussian estimator works well for smaller kernels, and in the literature, it has been shown that architectures with a hierarchy of smaller convolutions followed by nonlinearities can achieve equal (and often better) performance as those will fewer, larger kernels \citep{vgg16}. Thus, the ready-made Gaussian regularizer we introduced here can be used in many applications.

%For the purposes of our experiments in this paper, we used an early learning (or ``meta-learning") phase to develop the parameters of our CNN prior. Ultimately, we hope to discover general priors for each layer of visual processing that are effective across a wide variety of recognition tasks. This objective stands apart from the objectives of the meta-learning community. Meta-learning techniques provide a structure learning algorithm that must be applied to a large set of background training ``episodes" to yield structural principles. These principles may then later be applied in new tasks. In contrast, our goal is to develop a set of off-the-shelf structural principles that soft constraints AlexNet.... Analogy of the convolution... we could have ``learned" the convolution sparsity pattern by applying a fully-connected network to a massive image dataset... but instead we simply enforce the convolution sparsity

\section{Acknowledgements}
We thank Nikhil Parthasarathy and Emin Orhan for their valuable comments. Reuben Feinman is supported by a Google PhD Fellowship in Computational Neuroscience.

% bibliography

\bibliographystyle{apacite}

\setlength{\bibleftmargin}{.125in}
\setlength{\bibindent}{-\bibleftmargin}

\bibliography{ms}

\end{document}